\documentclass[runningheads]{llncs}

\usepackage[table,xcdraw]{xcolor}
\usepackage{multirow}

\usepackage[T1]{fontenc}
\usepackage{graphicx}

\usepackage{amsmath}
\usepackage{subfig}

\usepackage{todonotes}
\usepackage{comment}

\def\R{{\bbbr}}
\newcommand{\bigO}[1]{\ensuremath{\mathop{}\mathopen{}O\mathopen{}\left(#1\right)}}

\begin{document}

\title{A Self-Encoder for Learning Nearest Neighbors}

\author{Armand Boschin \and Thomas Bonald \and Marc Jeanmougin}
%

\institute{Télécom Paris, Institut Polytechniques de Paris, France \email{surname.name@telecom-paris.fr}}
\maketitle              
\begin{abstract}
We present the self-encoder, a neural network trained to guess the identity of each data sample. 
Despite its simplicity, it learns a very useful representation of data, in a self-supervised way. 
Specifically, the self-encoder learns to distribute the data samples in the embedding space so that they are linearly separable from one another.
This induces a geometry where two samples are close in the embedding space when they are not easy to differentiate. The self-encoder can then  be combined with a nearest-neighbor classifier or regressor for any subsequent supervised task.  Unlike regular nearest neighbors,   the predictions resulting from this encoding of data are invariant to any  scaling of  features, making any preprocessing like min-max scaling not necessary. The experiments show the efficiency of the approach, especially on heterogeneous data mixing numerical features and categorical features.
\end{abstract}

\section{Introduction} \label{sec:intro}

Despite the recent progress of machine learning, the question of the optimal encoding of data remains open, especially for tabular data \cite{borisov2022}.
In this paper, we present the self-encoder,  a neural network trained to guess the identity of each data sample.  Given $n$ data samples $x_1,\ldots,x_n\in \R^d$, the self-encoder maps any sample $x\in \R^d$ to a probability distribution $p$ over $\{1,\ldots,n\}$, in such a way that $p(x_i)$ is close to a Dirac in $i$ for each $i\in \{1,\ldots,n\}$.
In other words, the self-encoder is a classifier where each sample of the train set has its own label (its index). As such, it belongs to the category of self-supervised learning methods, like auto-encoders.
The key difference is that, while auto-encoders rely on a reconstruction task, with the output in the same space $\R^d$ as the original sample, our self-encoder relies on an identification task, with the output in the set of probability distributions on the  set of indices $\{1,\ldots, n\}$.


Despite its simplicity, the self-encoder learns a very useful representation of data. It learns to distribute data in the embedding space in a way that makes them  linearly separable from one another. This induces a geometry where two samples are close in the embedding space when they are not easy to differentiate.
In particular, the self-encoder can be used for any classification or regression task, using the $k$ nearest neighbors in the sense of this geometry, as given by the ranking of the predicted probabilities $p_1(x), \ldots, p_n(x)$, for any sample $x$. Interestingly, these nearest neighbors do not correspond to those given by the Euclidean distance in the original space $\R^d$ (nor by any other usual metric like a Minkowski metric or cosine similarity for instance). The geometry is {\it learned} by the model. In particular, the predictions resulting from this encoding of data are invariant to any  scaling of  the features, making any preprocessing like min-max scaling not necessary.

A  drawback of the self-encoder is its complexity, as the dimension of the output is equal to $n$, the size of the training set. This induces  a time complexity in $O(n^2)$ for training. 
To overcome this problem, we present a variant based on sampling where the model is trained to predict the identity of samples in a random subset of the training set, reducing the training time.

The rest of the paper is organized as follows. We present the related work in Section~\ref{sec:related}. The self-encoder is presented in Section~\ref{sec:encoder}. In Section~\ref{sec:invariance}, we prove that the learned geometry is invariant to  linear transformations, like any scaling of the features. The behavior of the self-encoder in the presence of categorical features is analyzed in Section~\ref{sec:categorical}. The variant of the model based on sampling is described in Section~\ref{sec:sampling}. The experiments are presented Section~\ref{sec:experiments}. Section~\ref{sec:conclusion}  concludes the paper.

\section{Related work} \label{sec:related}


\subsubsection{Nearest Neighbors.}
A simple and yet fundamental method to solve tasks in machine learning is the proximity search. It relies on the intuition that close vectors in the feature space should have close properties (similar labels in classification and close values in regression). The most common classifier is the $k$ nearest neighbor ($k$-NN) algorithm \cite{cover_nearest_1967}, which assigns a label to a point by choosing the most present label among its $k$ nearest neighbors. Usual similarity measures are the Euclidean distance or the cosine similarity but $k$-NN can also be applied to any similarity measure, such as that learned by the proposed self-encoder model.

\subsubsection{Kernel methods.} Many machine learning  algorithms are able to work with kernels, i.e.,  similarity functions that differ from the usual Euclidean metric. This is the case of the Support Vector Machine (SVM) \cite{cortes1995support}, a  classifier that tries to find linear  hyperplanes separating samples of different labels. When the dataset is not linearly separable, the SVM can still be used with the help of a kernel that moves the training vectors into another space, possibly of different dimension, in which they are more likely to be separable. 
The key point is that the kernel must be chosen by the user. In comparison, the self-encoder {\it learns} a geometry that makes data samples linearly separable from another. Moreover, it is a self-supervised learning technique, that does not rely on labeled data. 

\subsubsection{Auto-encoders.}
In its simplest form, an auto-encoder is a neural network model that learns a compact latent representation of data points with an encoder part and that is able to reconstruct them from this representation with a decoder part. The rationale is that a performing encoder should be able to extract the core discriminative features that define the samples of the dataset and this is measured by the ability of the decoder to reconstruct the original data from those features.

There is a large variety of application domains for auto-encoders and even more different structures. Let us cite for example variational auto-encoders that are used in a probabilistic framework for variational inference \cite{kingma2014auto} or denoising auto-encoders that can be used for example to improve image resolutions by changing the reconstruction criterion \cite{vincent_stacked_2010} of manually noised data samples.

The output of an auto-encoder being a reconstructed version of the input vector, the measure of reconstruction loss is key in the training process. Unlike auto-encoders, in the case of the Self-Encoder, there is no need to engineer a good similarity measure between the original data sample and its reconstruction because the objective is simply to predict the identity of each data sample as an index in $\{1,\ldots,n\}$ and not the data point in $\R^d$ itself.

\section{The self-encoder} \label{sec:encoder}

Let $x_1,\ldots,x_n \in \R^d$ be the set of training samples with $d$ the dimension of the feature space and $n$ the number of samples. 
The Self-Encoder is a multi-layer perceptron with input dimension $d$ and output dimension $n$, trained to predict the identity $i$ of each data sample $x_i$.

\subsubsection{Hidden layers} 
The encoder consists of $L$ hidden layers. Each layer $l=1,\ldots,L$ consists in an affine transformation followed by 
an activation function:
\begin{align*}
&h^{(1)} = \phi^{(1)}\left(W^{(1)}x + b^{(1)}\right)\\
&h^{(2)} = \phi^{(2)}\left(W^{(2)}h^{(1)} + b^{(2)}\right)\\
&\vdots\\
&h^{(L)}= \phi^{(L)}\left(W^{(L)}h^{(L-1)} + b^{(L)}\right)
\end{align*}
where $\phi^{(1)},\ldots,\phi^{(L)}$ are the activation functions, typically non-linear.
The dimensions of the successive outputs, say $d_1,\ldots,d_L$, are hyper-parameters. 
 The weight matrices $W^{(1)},\ldots, W^{(L)}$ and the bias vectors $b^{(1)},\ldots,b^{(L)}$ must be learned.

\subsubsection{Output layer}
The output layer is a fully connected layer with input dimension $d_L$ (the output dimension of the last hidden layer) and output dimension $n$. This affine transformation is followed by an activation function $\phi$ which is either a coordinate-wise sigmoid function:
$$\forall u \in \R^n,\quad \phi(u) = \left(\frac {e^{u_1}} {1+e^{u_1}},\ldots, \frac {e^{u_n}} {1+e^{u_n}}\right)$$
or a SoftMax function:
$$\forall u \in \R^n,\quad \phi(u) = \left(\frac {e^{u_1}} {\sum_{i=1}^n e^{u_i}},\ldots, \frac {e^{u_n}} {\sum_{i=1}^n e^{u_i}}\right)$$

The output of the network is then a vector $p = \phi(W h^{(L)} + b) \in[0, 1]^n$ that can be interpreted as probabilities: 
the $i$th component $p_i$ is the probability that the input $x$ corresponds to the training sample $x_i$. 
Observe that the probabilities sum to 1 only with the SoftMax function (with the sigmoid function, the probabilities are learned independently for each training sample $i$ by the output layer).
The weight matrix $W$ and the bias vector $b$ must be learned, together with the other parameters.

\subsubsection{Loss function}
In the following, $f$ denotes the learned function of the network, mapping sample vectors $x\in \R^d$ to probability vectors $p\in[0,1]^n$. $f$ is a parametric function and its parameters are the weight matrices $W^{(1)},\ldots, W^{(L)}, W$ and the bias vectors $b^{(1)},\ldots,b^{(L)}, b$. Those are learned by minimizing the following Binary Cross Entropy (BCE) loss by gradient descent:
\begin{equation}
\label{eq:wit_bce}
{\cal L} = - \sum_{i=1}^n \left( \log f_i(x_i) +\sum_{j\ne i} \log(1 - f_j(x_i))\right) 
\end{equation}

\subsubsection{Interpretation}
The Self-Encoder learns a latent representation of data, given by the last hidden layer, where the $n$ training samples are linearly separable.
It is the role of the output layer to find the hyperplanes (given by the weight matrix $W$ and the bias vector $b$) separating each training sample. 

\subsubsection{No hidden layer}
Observe that the Self-Encoder can also be trained without any hidden layer. It then reduces to the output layer, i.e. a perceptron with input dimension $d$ and output dimension $n$. Equivalently, the Self-Encoder then consists of $n$ binary logistic regressions (for the sigmoid activation function) or a single multinomial logistic regression (for the SoftMax activation function).

\subsubsection{Geometry}
In both cases (with or without hidden layers), the Self-Encoder learns a specific similarity measure in the sense that it can predict the training samples that are the most similar to any new data sample $x$.
This measure depends on the distribution of the training samples $x_1,\ldots,x_n$ in the original space $\R^d$. Any sample $x\in \R^d$ is said to be {\it close} to the training sample $x_i$ if the corresponding predicted probability $p_i = f_i(x)$ is close to 1.

\section{Invariance property} \label{sec:invariance}

The Self-Encoder is invariant to invertible affine transformations of the training data, as stated below.

\begin{proposition}\label{prop:invariance}
Let $f$ be the mapping learned by the encoder with training samples $x_1,\ldots, x_n$. For any invertible matrix $M\in \R^{d\times d}$ and vector $v\in \R^d$, let
$\tilde {x}_1,\ldots, \tilde {x}_n$ be the new training samples obtained by affine transformation $x\mapsto Mx + v$. The new mapping $\tilde f$ defined by:
$$
\forall \tilde {x} \in \R^d,\quad 
\tilde f(\tilde {x}) = f(M^{-1}(\tilde {x}-v))
$$
minimizes the cross-entropy loss \eqref{eq:wit_bce} for the training samples $\tilde {x}_1,\ldots, \tilde {x}_n$. Both encoders are related through the affine transformation:
$$
\forall x\in \R^d,\quad f(x) = \tilde f(Mx + v).
$$
\end{proposition}

In other words, if the training data are the same up to some invertible affine transformations, so are the mappings learned by the encoder.

\begin{proof}
The mapping $\tilde f$ is the same encoder as $f$ except for the first hidden layer, whose output $\tilde {h}^{(1)}$ for the input $\tilde {x}$ is given by:
\begin{align*}
    \tilde {h}^{(1)}\left(\tilde {x}\right) &= {h}^{(1)}\left(M^{-1}(\tilde {x}-v)\right)\\
    &= \phi^{(1)}\left(W^{(1)}(M^{-1}(\tilde {x}-v)) + b^{(1)}\right)\\
    &= \phi^{(1)}\left(W^{(1)}M^{-1}\tilde {x} + b^{(1)} - W^{(1)} M^{-1} v\right)\\
    & = \phi^{(1)}(\tilde W^{(1)}\tilde {x} + \tilde {b}^{(1)}),
\end{align*}
for the weight matrix and bias vector:
\begin{align*}
\tilde W^{(1)} &= W^{(1)}M^{-1}\\
\tilde {b}^{(1)} &= b^{(1)} - W^{(1)} M^{-1} v.
\end{align*}
The corresponding binary cross-entropy loss is minimized, given that:
\begin{align*}
{\cal L} &= - \sum_{i=1}^n \left( \log \tilde f_i(\tilde{x}_i) +\sum_{j\ne i} \log(1 - \tilde f_j(\tilde{x}_i))\right)\\
&= - \sum_{i=1}^n \left( \log f_i(x_i) +\sum_{j\ne i} \log(1 - f_j( x_i))\right)
\end{align*}
\end{proof}

\begin{figure}
    \centering
    \subfloat[Euclidean distance]{
        \resizebox{0.3\columnwidth}{!}{
        \begin{tikzpicture}[thick]
            \draw[] (-2,-0) -- (2,0);
            \draw[] (0,-2) -- (0,2);
            \fill (1,1) circle (0.1);
            \fill (1,-1) circle (0.1);
            \fill (-1,1) circle (0.1);
            \fill (-1,-1) circle (0.1);
        \end{tikzpicture}}
        \hspace{1cm}
        \resizebox{0.4\columnwidth}{!}{
            \begin{tikzpicture}[thick]
                \draw[] (-0.55,-0.55) -- (0.55,0.55);
                \draw[] (0.55,0.55) -- (2.5,-0.7);
                \draw[] (-0.55,-0.55) -- (-2.5,0.7);
                \draw[] (-0.55,-0.55) -- (-0.2,-2);
                \draw[] (0.55,0.55) -- (0.2,2);
                \fill (1.5,1) circle (0.1);
                \fill (0.5,-0.5) circle (0.1);
                \fill (-1.5,-1) circle (0.1);
                \fill (-0.5,0.5) circle (0.1);
            \end{tikzpicture}
        }
    }
    
    \subfloat[Self-Encoder]{
        \resizebox{0.3\columnwidth}{!}{
            \begin{tikzpicture}[thick]
                \draw[] (-2,-0) -- (2,0);
                \draw[] (0,-2) -- (0,2);
                \fill (1,1) circle (0.1);
                \fill (1,-1) circle (0.1);
                \fill (-1,1) circle (0.1);
                \fill (-1,-1) circle (0.1);
            \end{tikzpicture}
        }
        \hspace{1cm}
        \resizebox{0.4\columnwidth}{!}{
            \begin{tikzpicture}[thick]
                \draw[] (-3,-0.75) -- (3,0.75);
                \draw[] (1.33,2) -- (-1.33,-2);
                \fill (1.5,1) circle (0.1);
                \fill (0.5,-0.5) circle (0.1);
                \fill (-1.5,-1) circle (0.1);
                \fill (-0.5,0.5) circle (0.1);
            \end{tikzpicture}
        }
    }

    \centering
    \caption{Impact of a linear transformation on the Voronoi diagram (regions of nearest neighbors) formed by $n=4$ points in $\R^2$.}
    \label{fig:voronoi}
\end{figure}
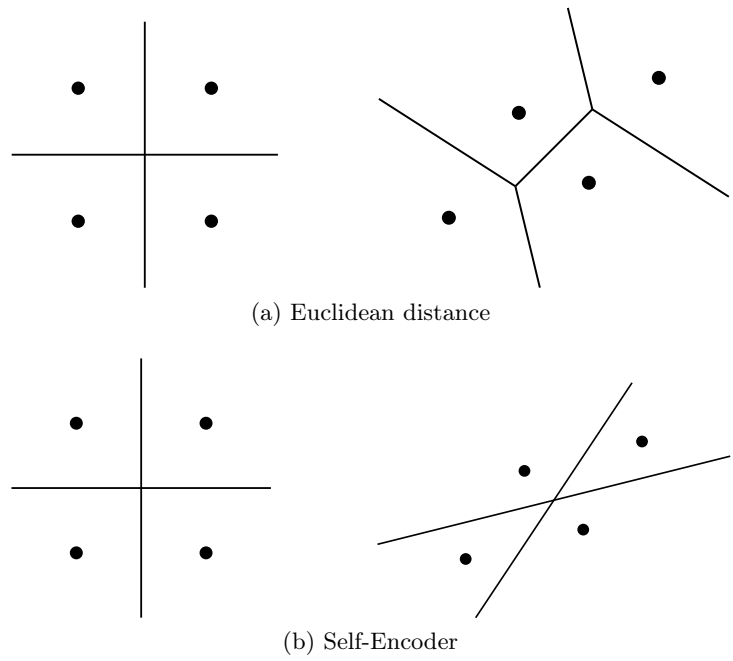

This invariance property is a clear difference with the usual Euclidean distance. To illustrate this, Figure \ref{fig:voronoi} shows the Voronoi diagram (regions formed by the nearest neighbors) associated with $n=4$ points of $\R^2$ for the Self-Encoder and for the Euclidean distance. On the left, the 4 points form a square and the Voronoi diagrams coincide, by symmetry. On the right, a linear transformation is applied; the Voronoi diagram is obtained by the same linear transformation for the Self-Encoder, while it changes completely for the Euclidean distance.

This invariance property is handy as it simplifies the pre-processing steps, which can become tedious and that usually have an impact on the performances of many classifiers. This will be showed later in the experiments for the Euclidean $k$-NN classifier.

\section{Categorical features} \label{sec:categorical}

We claim that the Self-Encoder robustly handles categorical features, in the sense that it does not depend on the number of bits they are encoded on, unlike Euclidean Nearest Neighbor (NN).

Given $x^{(1)}, \dots, x^{(n)}$ training samples with categorical features in $\{0, 1\}^d$, data redundancy in the features does not modify the optimum of the function. If there is a pair $(k_1, k_2)$ such that $\forall i, x^{(i)}_{k_1} = x^{(i)}_{k_2}$ then the contributions to the loss involving these features is duplicated and they should result in the same corresponding weights in the input layer, which should not impact the learned geometry. If there is a pair $(k_1, k_2)$ such that $\forall i, x^{(i)}_{k_1} = 1 - x^{(i)}_{k_2}$ then the $k_1$-th feature is a invertible affine transformation of the $k_2$-th feature and the invariance property of Proposition~\ref{prop:invariance} tells us that the first case of redundancy applies.

Let us illustrate on a simple example how the similarity measure learned by a Self-Encoder can outperform the usual Euclidean metric on a nearest-neighbor search by handling categorical features differently.
\begin{align}
X_1 &=\begin{pmatrix} x_1^{(1)} \\ x_1^{(2)} \\ x_1^{(3)} \\ x_1^{(4)} \\x_1^{(5)}\end{pmatrix} = \begin{pmatrix}0 & 0 & 0 & 1\\0 & 0 & 1 & 0\\0 & 1 & 0 & 0\\ 1 & 0 & 0 & 1\\ 1 & 0 & 1 & 0\\\end{pmatrix} \label{eq:matrices1}\\
X_2 &=\begin{pmatrix} x_2^{(1)} \\ x_2^{(2)} \\ x_2^{(3)} \\ x_2^{(4)} \\x_2^{(5)}\end{pmatrix} = \begin{pmatrix}
1 & 0 & 0 & 0 & 1\\1 & 0 & 0 & 1 & 0\\1 & 0 & 1 & 0 & 0\\0 & 1 & 0 & 0 & 1\\ 0 & 1 & 0 & 1 & 0\\\end{pmatrix}\label{eq:matrices2}
\end{align}

Let $X_1$ and $X_2$ be defined in Equations~\ref{eq:matrices1} and \ref{eq:matrices2}. The only difference between the two set of samples is that the binary feature of the first binary column of $X_1$ is encoded on two bits in $X_2$ in the first two columns. It is reasonable to expect that a classifier trained on $X_1$ and fed with $\bar {x}_1 = \begin{pmatrix}1 & 1 & 0 & 0 \end{pmatrix}$ makes the same decision as a classifier trained on $X_2$ and fed with $\bar {x}_2 = \begin{pmatrix}0 & 1 & 1 & 0 & 0 \end{pmatrix}$. This is not the case for a Euclidean nearest-neighbor as the closest vector from $\bar {x}_1$ in $X_1$ is $\begin{pmatrix}0 & 1 & 0 & 0 \end{pmatrix}$ and the closest ones from $\bar {x}_2$ in $X_2$ are $\begin{pmatrix}1 & 0 & 1 & 0 & 0 \end{pmatrix}$, $\begin{pmatrix} 0 & 1 & 0 & 0 & 1 \end{pmatrix}$ and $\begin{pmatrix}0 & 1 & 0 & 1 & 0 \end{pmatrix}$. On the other hand, the Self-Encoder
returns $\begin{pmatrix}0 & 1 & 0 & 0 \end{pmatrix}$ as the most similar to $\bar{ x}_1$ and $\begin{pmatrix}1 & 0 & 1 & 0 & 0 \end{pmatrix}$ as the most similar to $\bar {x}_2$.

\section{Sampling} \label{sec:sampling}

Under the reasonable assumption that the dimensions of the hidden layers do not scale with $n$, the time complexity is that of the output layer: $O(n^2)$ for training and $O(n)$ for evaluation. The training time complexity is 
higher than the $n$-linear time complexity of Euclidean $k$-NN. However, training is only done once and then the trained model can be used in $\bigO{n}$ for similarity search. Moreover, the natural way to implement a Multi-Layer Perceptron (MLP) model is to use machine learning frameworks such as PyTorch \cite{pytorch} or TensorFlow \cite{tensorflow2015-whitepaper}, which natively support GPU acceleration and highly reduce the computation time.

To control the memory usage and time complexity of the approach and in order to improve the performance, we propose to use a simple sampling strategy, where a random subset of the training set is selected, thus reducing the space and time complexity of the model. Given a set of training samples $X = (x^{(i)})_{i\in \{1,\dots,n\}}$, sampling generates a new training subsets of size $s$ by randomly sampling vectors from $X$. The model is then trained on the new training subset. Experiments show that that the performance of the sampled model is comparable to the complete one.

\section{Experiments}\label{sec:experiments}

To evaluate the quality of the similarity measure learned by the proposed Self-Encoder, we apply it to a  classification task with $k$-NN method . The Self-Encoder $k$-NN is simply called Self-Encoder for simplicity. It is tested with and without a hidden layer and in normal and sampling modes. Training uses early stopping and learning rate decay.

The performances are compared to those of four classification baselines: the Euclidean $k$-NN with $k=5$, the linear Support Vector Machine (SVM), the one-vs-all logistic regression and the multi-layer perceptron with one hidden layer of twenty neurons.

Ten datasets that are recurrent in the machine learning literature were selected for the experiments. They are all available from the UCI repository\footnote{\url{https://archive.ics.uci.edu/ml/datasets.php}} and descriptive figures can be found in Table~\ref{tab:wit_datasets}. Among these, the German Credit dataset comes in two versions: one with numerical features and another one with 13 out of 20 features being categorical. The only pre-processing applied to all the datasets is the conversion of categorical features into a one-hot encoding.

\begin{table}
\centering
\begin{tabular}{l|l|l|l|}
\cline{2-4}
                                                                                            & \# samples & \begin{tabular}[c]{@{}l@{}}Feature \\ dimension\end{tabular} & \# classes \\ \hline
\multicolumn{1}{|l|}{Breast Cancer} & 699 & 9 & 2 \\ \hline
\multicolumn{1}{|l|}{Digits} & 1,797 & 64 & 10 \\ \hline
\multicolumn{1}{|l|}{Ecoli} & 336 & 7 & 8 \\ \hline
\multicolumn{1}{|l|}{German credit} & 1,000 & 24 & 2 \\ \hline
\multicolumn{1}{|l|}{\begin{tabular}[c]{@{}l@{}}German credit\\ (categorical)\end{tabular}} & 1,000 & 20 & 2 \\ \hline
\multicolumn{1}{|l|}{Glass} & 214 & 9 & 6 \\ \hline
\multicolumn{1}{|l|}{Ionosphere} & 351 & 34 & 2 \\ \hline
\multicolumn{1}{|l|}{Iris} & 150 & 4 & 3 \\ \hline
\multicolumn{1}{|l|}{Liver} & 345 & 6 & 2 \\ \hline
\multicolumn{1}{|l|}{Wine} & 178 & 13 & 3 \\ \hline
\end{tabular}
\caption{Dataset details}
\label{tab:wit_datasets}
\end{table}

The Self-Encoder is implemented in PyTorch, the optimization is done using Adam \cite{adam}. Some hyper-parameters are fixed: the learning rate decay is set to 0.995, the size of the hidden layer is fixed to 20 and for the sampling mode, the number of visible samples is fixed to 100. 

For each classification model and each dataset, the reported metric is the accuracy. It is measured using 5-fold cross validation. The learning rate to train each model on each \textit{fold} is chosen according to a log-uniform distribution between $0.001$ and $2$, using the Bayesian optimization library \texttt{hyperopt}\footnote{\url{https://hyperopt.github.io/hyperopt}}. The library is also used to choose the normalization function between sigmoid and SoftMax. All experiments can be reproduced using the code provided in the supplementary material.

\begin{table}
    \centering
    \subfloat[Classification accuracy of the baseline methods.]{
        \begin{tabular}{l|ll|ll|ll|ll|}
            \cline{2-9}
                             & \multicolumn{2}{l|}{\multirow{2}{*}{$k$-NN}} & \multicolumn{2}{l|}{\multirow{2}{*}{MLP}} & \multicolumn{2}{l|}{\multirow{2}{*}{Logistic}} & \multicolumn{2}{l|}{\multirow{2}{*}{SVM}} \\
                             & \multicolumn{2}{l|}{}                        & \multicolumn{2}{l|}{}                     & \multicolumn{2}{l|}{}                          & \multicolumn{2}{l|}{} \\ \hline
            \multicolumn{1}{|l|}{Breast cancer} & \textbf{0.967} & 0.009 & 0.96 & 0.007 & 0.963 & 0.003 & 0.96 & 0.003  \\ \hline
            \multicolumn{1}{|l|}{Digits}        & \textbf{0.983} & 0.004 & 0.934 & 0.013 & 0.965 & 0.004 & \textbf{0.983} & 0.002 \\ \hline
            \multicolumn{1}{|l|}{Ecoli}         & \textbf{0.869} & 0.021 & 0.762 & 0.016 & 0.759 & 0.029 & 0.798 & 0.019  \\ \hline
            \multicolumn{1}{|l|}{German credit} & 0.684 & 0.012 & 0.741 & 0.026 & 0.738 & 0.012 & \textbf{0.754} & 0.012  \\ \hline
            \multicolumn{1}{|l|}{German credit cat} & 0.716 & 0.027 & 0.717 & 0.023 & 0.733 & 0.027 & \textbf{0.736} & 0.023  \\ \hline
            \multicolumn{1}{|l|}{Glass}         & \textbf{0.616} & 0.066 & 0.439 & 0.099 & 0.579 & 0.046 & \textbf{0.616} & 0.06  \\ \hline
            \multicolumn{1}{|l|}{Ionosphere}    & 0.835 & 0.032 & \textbf{0.903} & 0.033 & 0.855 & 0.039 & 0.838 & 0.03  \\ \hline
            \multicolumn{1}{|l|}{Iris}          & 0.953 & 0.05 & 0.913 & 0.129 & 0.967 & 0.03 & \textbf{0.987} & 0.027  \\ \hline
            \multicolumn{1}{|l|}{Liver}         & \textbf{0.693} & 0.025 & 0.609 & 0.098 & 0.684 & 0.063 & 0.687 & 0.057 \\ \hline
            \multicolumn{1}{|l|}{Wine}          & 0.714 & 0.046 & 0.321 & 0.1 & 0.966 & 0.021 & \textbf{0.977} & 0.011 \\ \hline
        \end{tabular}
        }
    
    \subfloat[Classification accuracy of the self-encoder. The score of the best baseline is recalled for comparison.]{
        \begin{tabular}{l|ll|ll|ll|}
            \cline{2-7}
                & \multicolumn{2}{l|}{\multirow{2}{*}{SE}} & \multicolumn{2}{l|}{\multirow{2}{*}{SE hidden}} & \multicolumn{2}{l|}{\multirow{2}{*}{Best Baseline}}\\
                & \multicolumn{2}{l|}{} & \multicolumn{2}{l|}{} & \multicolumn{2}{l|}{} \\ \hline
            \multicolumn{1}{|l|}{Breast cancer} & 0.969 & 0.007 & \textbf{0.976} & 0.011 & 0.967 & 0.009 \\ \hline
            \multicolumn{1}{|l|}{Digits}        & \textbf{0.983} & 0.012 & 0.655 & 0.117 &\textbf{0.983} & 0.004 \\ \hline
            \multicolumn{1}{|l|}{Ecoli}         & \textbf{0.911} & 0.028 & 0.899 & 0.021 & 0.869 & 0.021 \\ \hline
            \multicolumn{1}{|l|}{German credit} & \textbf{0.768} & 0.011 & 0.749 & 0.007 & 0.754 & 0.012 \\ \hline
            \multicolumn{1}{|l|}{German credit cat} & \textbf{0.768} & 0.025 & 0.756 & 0.02 & 0.736 & 0.023 \\ \hline
            \multicolumn{1}{|l|}{Glass}         & \textbf{0.794} & 0.033 & 0.766 & 0.041 & 0.616 & 0.06 \\ \hline
            \multicolumn{1}{|l|}{Ionosphere}    & 0.94 & 0.023 & \textbf{0.963} & 0.031 & 0.903 & 0.033 \\ \hline
            \multicolumn{1}{|l|}{Iris}          & \textbf{1.0} & 0.0 & 0.993 & 0.013 & 0.987 & 0.027 \\ \hline
            \multicolumn{1}{|l|}{Liver}         & 0.759 & 0.02 & \textbf{0.768} & 0.016 & 0.693 & 0.025 \\ \hline
            \multicolumn{1}{|l|}{Wine}          &  0.736 & 0.071 & 0.871 & 0.116 & \textbf{0.977} & 0.011 \\ \hline
        \end{tabular}
        }

    \subfloat[Classification accuracies of the Self-Encoder in sampling mode (with 100 visible samples chosen uniformly at random). The score of the best baseline is recalled for comparison.\label{tab:res_sampling}]{
        \begin{tabular}{l|ll|ll|ll|}
            \cline{2-7}
                & \multicolumn{2}{l|}{\multirow{2}{*}{SE}} & \multicolumn{2}{l|}{\multirow{2}{*}{SE hidden}} & \multicolumn{2}{l|}{\multirow{2}{*}{Best Baseline}}\\
                & \multicolumn{2}{l|}{} & \multicolumn{2}{l|}{} & \multicolumn{2}{l|}{} \\ \hline
            \multicolumn{1}{|l|}{Breast cancer} & 0.960 & 0.006 & \textbf{0.970} & 0.008 & 0.967 & 0.009 \\ \hline
            \multicolumn{1}{|l|}{Digits}        & 0.853 & 0.019 & 0.528 & 0.161 &\textbf{0.983} & 0.004 \\ \hline
            \multicolumn{1}{|l|}{Ecoli}         & 0.878 & 0.036 & \textbf{0.899} & 0.025 & 0.869 & 0.021 \\ \hline
            \multicolumn{1}{|l|}{German credit} & 0.741 & 0.012 & 0.750 & 0.008 & \textbf{0.754} & 0.012 \\ \hline
            \multicolumn{1}{|l|}{German credit cat} & 0.742 & 0.011 & \textbf{0.750} & 0.011 & 0.736 & 0.023 \\ \hline
            \multicolumn{1}{|l|}{Glass}         & 0.766 & 0.03 & \textbf{0.771} & 0.027 & 0.616 & 0.06 \\ \hline
            \multicolumn{1}{|l|}{Ionosphere}    & 0.897 & 0.028 & \textbf{0.946} & 0.017 & 0.903 & 0.033 \\ \hline
            \multicolumn{1}{|l|}{Iris}          & \textbf{1.0} & 0.0 & 0.993 & 0.013 & 0.987 & 0.027 \\ \hline
            \multicolumn{1}{|l|}{Liver}         & \textbf{0.762} & 0.024 & 0.760 & 0.007 & 0.693 & 0.025 \\ \hline
            \multicolumn{1}{|l|}{Wine}          & 0.702 & 0.04 & 0.848 & 0.073 & \textbf{0.977} & 0.011 \\ \hline
        \end{tabular}
        }
    \caption{Classification accuracies for the Self-Encoder along with other baselines on ten usual datasets. Performances are measured using a 5-fold cross validation mechanism. Best results are in bold. German credit cat is the version of the German credit dataset with categorical features.} 
    \label{tab:res_wit}
\end{table}
\clearpage

All results are reported in Table~\ref{tab:res_wit}. An obvious conclusion is that in both normal and sampling settings, the Self-Encoder performs better than the baselines. A second conclusion is that the Self-Encoder in the sampling framework performs comparably or better than the other baselines in the normal setting. This is reassuring because the Self-Encoder might need to be applied in sampling mode while other lighter models have access to all the samples.

To measure the impact of the invariance feature, the performance of the Self-Encoder has also been compared to the performance of the Euclidean $k$-NN algorithm on normalized numerical datasets. The results are shown in Table~\ref{tab:normalized_datasets}. As expected, normalization improves the score of the Euclidean $k$-NN in most cases but not up to the score of the Self-Encoder.

\begin{table}
    \centering
    \begin{tabular}{c|c|c|c|}
    \cline{2-4}
& \begin{tabular}[c]{@{}c@{}}$k$-NN \\ normalized\end{tabular} & $k$-NN & SE \\ \hline
\multicolumn{1}{|l|}{Breast cancer} & 0.960 & \textbf{0.967} & \textbf{0.976} \\ \hline
\multicolumn{1}{|l|}{Digits} & 0.970 & \textbf{0.983} & \textbf{0.983} \\ \hline
\multicolumn{1}{|l|}{Ecoli} & 0.866 & \textbf{0.869} & \textbf{0.911} \\ \hline
\multicolumn{1}{|l|}{German} & \textbf{0.693} & 0.684 & \textbf{0.768} \\ \hline
\multicolumn{1}{|l|}{Glass} & \textbf{0.645} & 0.616 & \textbf{0.794} \\ \hline
\multicolumn{1}{|l|}{Ionosphere} & \textbf{0.835} & \textbf{0.835} & \textbf{0.963} \\ \hline
\multicolumn{1}{|l|}{Iris} & \textbf{0.960} & 0.953 & \textbf{1.0} \\ \hline
\multicolumn{1}{|l|}{Liver} & 0.609 & \textbf{0.693} & \textbf{0.768} \\ \hline
\multicolumn{1}{|l|}{Wine} & \textbf{0.966} & 0.714 & \textbf{0.977} \\ \hline
   \end{tabular}
    \caption{Classification accuracy of Euclidean $k$-NN on normalized datasets and Euclidean $k$-NN and Self-Encoder on raw datasets. This comparison is only reported on numerical datasets. Two best scores for each dataset are in bold.}
    \label{tab:normalized_datasets}
\end{table}

\section{Conclusion}
\label{sec:conclusion}

In conclusion, the Self-Encoder is an unsupervised method that learns a similarity measure specific to the training data that can be used for downstream tasks such as classification, its primary objective being to separate samples from one another.

\newpage
\bibliographystyle{splncs04}
\bibliography{biblio}

\end{document}